\documentclass[]{bytedance_seed}

\usepackage[toc,page,header]{appendix}
\usepackage{minitoc}
\usepackage[utf8]{inputenc}
\usepackage[ruled,linesnumbered]{algorithm2e}
\usepackage[noend]{algpseudocode}
\usepackage{amsmath,amsthm,amsfonts,amssymb}
\usepackage{hyperref}
\usepackage{color}
\usepackage{mathrsfs}
\usepackage{enumitem}
\usepackage{multirow}
\usepackage{booktabs}
\usepackage{makecell}
\usepackage{graphicx}
\usepackage{caption}
\usepackage{thmtools}
\usepackage{thm-restate}
\usepackage{hhline}
\usepackage{cite}
\usepackage[table]{xcolor}
\usepackage{rotating} 
\usepackage{natbib}
\usepackage{bm}
\usepackage{subcaption}

\usepackage{tcolorbox}

\usepackage{tikz}
\usetikzlibrary{shapes,snakes}

\renewcommand{\epsilon}{\varepsilon}

\newcommand{\cA}{\mathcal{A}}

\newcommand{\cO}{\mathcal{O}}

\newcommand{\cR}{\mathcal{R}}
\newcommand{\cS}{\mathcal{S}}

\newcommand{\mP}{{\mathbb{P}}}

\theoremstyle{definition}

\newcount\Comments  %
\newcommand{\kibitz}[2]{\ifnum\Comments=1\textcolor{#1}{#2}\fi}

\newcommand{\blue}[1]{{\color{blue} #1}}

\tcbset{
  promptbox/.style={
    colback=blue!5!white,        %
    colframe=black!50,           %
    coltitle=black,              %
    colbacktitle=blue!10,        %
    fonttitle=\bfseries,
    boxrule=0.4pt,
    arc=1mm,
    left=1mm,
    right=1mm,
    top=1mm,
    bottom=1mm,
    title={},
  }
}

\tcbset{
  rubricsbox/.style={
    colback=red!5!white,        %
    colframe=black!50,           %
    coltitle=black,              %
    colbacktitle=red!10,        %
    fonttitle=\bfseries,
    boxrule=0.4pt,
    arc=1mm,
    left=1mm,
    right=1mm,
    top=1mm,
    bottom=1mm,
    title={},
  }
}

\title{ Beyond Verifiable Rewards: Rubric-Based GRM for Reinforced Fine-Tuning SWE Agents }

\author[1,*]{Jiawei Huang}
\author[2]{Qingping Yang}
\author[2]{Renjie Zheng}
\author[2]{Jiaze Chen}

\affiliation[1]{ETH Zurich}
\affiliation[2]{ByteDance Seed}

\contribution[*]{Work done at ByteDance Seed}

\abstract{
    Despite recent progress in Large Language Model (LLM) Agents for Software Engineering (SWE) tasks, end-to-end fine-tuning typically relies on verifiable terminal rewards such as whether all unit tests pass.
    While these binary signals reflect whether the final solution is correct, they provide little guidance for shaping intermediate behaviors during multi-step interactions, thereby limiting improvements in the overall quality of the resolution process.
    To address this, we introduce a rubric-based Generative Reward Model (GRM) that provides richer learning signals. 
    The GRM is equipped with human-designed rubrics that indicate criteria for encouraging or discouraging specific behavioral patterns, and we leverage this feedback for high-quality training data collection via trajectory filtration. 
    When used for Reinforced Fine-Tuning (RFT) on SWE Tasks, our approach outperforms terminal-score-only rejection sampling: it more effectively suppresses undesirable patterns while promoting beneficial ones, as confirmed by case analyses, and it ultimately improves final test accuracy.
}

\date{\today}
\correspondence{Jiawei Huang, \email{jiawei.huang@inf.ethz.ch}}

\begin{document}
\maketitle

\section{Introduction}

Large Language Model (LLM) Agents extend single-turn formulation by enabling multi-step interaction with environments \citep{wang2024survey}.
Rather than producing a static response, agents can invoke external tools calling, gather observations, and iteratively refine partial solutions, which are capabilities essential for long-horizon reasoning and decision making.
A particularly impactful instantiation of this paradigm is the coding agents \citep{yang2025qwen3}: by combining an LLM with tools such as code execution, unit testing, and documentation retrieval, they can automatically write, debug, and refine software.

A milestone in this direction is SWE-agent \citep{yang2024swe}, which demonstrates that LLM agents can solve real-world bug-fixing tasks in large codebases, having potential to integrate seamlessly with developer workflows and significantly reducing the manual effort required in software development.
While prompt engineering and scaffolding matter, further gains typically require fine-tuning the underlying LLM to enhance their problem-solving capabilities.
Analogous to math \citep{shao2024deepseekmath} or code generation \citep{gehring2024rlef}, a key advantage of SWE tasks is the availability of \emph{verifiable reward}: a binary signal indicating whether the final patch passes all unit tests or not---serves as an ideal learning signal directly reflecting the task succeeds or not.
Notably, such rewards are automatically checkable without human intervention, thereby making them attractive training signals.
As demonstrated by previous works \citep{pan2412training}, leveraging those reward signals for post-training can substantially improve the success rate.

\begin{figure}
    \centering
    \includegraphics[scale=0.48]{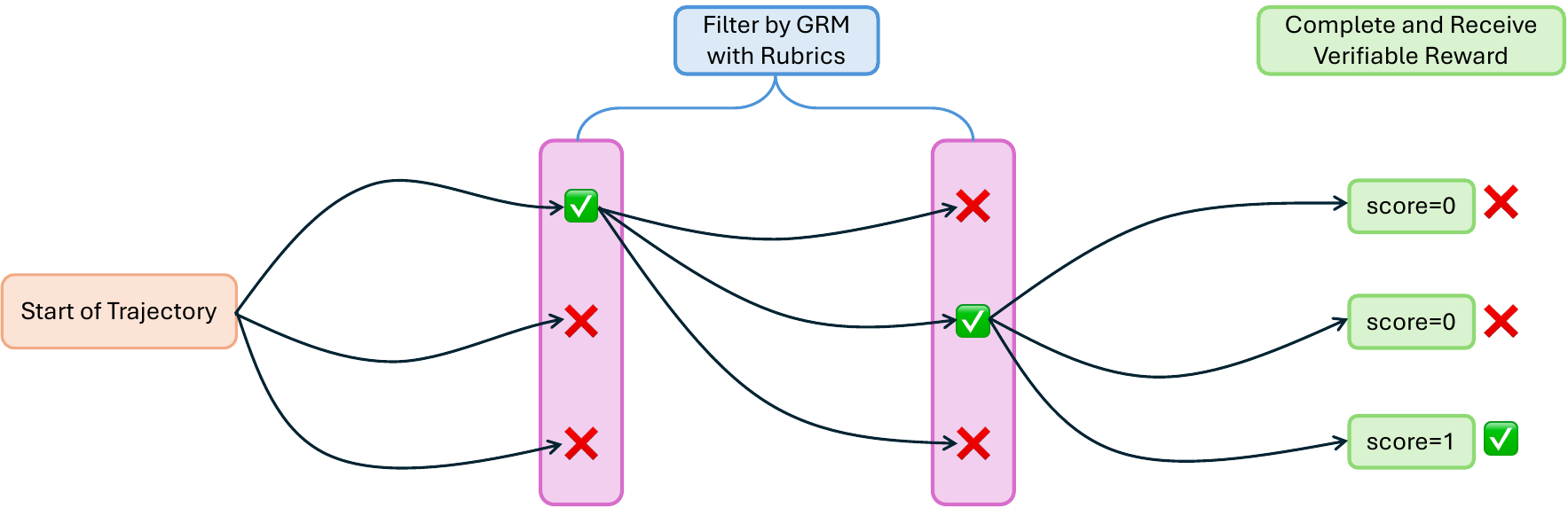}
    \caption{\textbf{Illustration of Rubrics-based GRM Filtering Process.} 
    From the same trajectory prefix (conversation history), we sample multiple candidates continuations (black arrows, three candidates per-step in figure). Here each candidate can be a one-step action or a multistep trajectory segments, which are sampled independently. 
    A Generative Reward Model (GRM) then evaluates candidates based on the rubrics and additional information when available (e.g., a ground-truth git patch), selects out the best continuation, and discards the others. This process repeats until the termination of the trajectory, after which a verifiable reward based on unit tests is assigned.}\label{fig:workflow}
\end{figure}

However, although such verifiable rewards accurately reflect whether the final solution works, they offer little guidance on the quality of intermediate reasoning steps, partial solutions, or error-handling strategies.
This becomes problematic when we care not only about pass rates, but also about the quality and efficiency of the resolution process. 
Here we highlight two concrete failure modes of training with terminal verifiable rewards only:
\begin{itemize}[leftmargin=*]
    \item \textbf{Persistence of self-correctable bad patterns} Training solely with verifiable rewards can not eliminate inefficient intermediate steps or erroneous actions that are later self-corrected by the agent.
    In SWE-tasks, for instance, agents may repeatedly invoke the same tool, or trigger recoverable execution errors (e.g., referencing a non-existent file or introducing temporary bugs that are fixed later). Because these behaviors do not necessarily reduce the probability of eventual success, they are not effectively visible to binary unit-test signals.
    The long-horizon nature in coding tasks further amplifies this issue: there is ample opportunity for inefficiencies to accumulate without changing the final outcome.
    
    \item \textbf{Lack of credit for beneficial intermediate patterns} Conversely, many intermediate behaviors, even if not significantly increase the probability of passing all tests, are valuable because their potential in improving efficiency and generalization ability if reinforced.
    Examples include writing a minimal reproduction script, or creating and running unit tests.
    Under terminal-only supervision, trajectories that exhibit these desirable patterns but fail due to a minor final mistake receive the same zero reward as completely unstructured attempts.
    As a result, solely optimizing terminal rewards risks overlooking and underutilizing such beneficial behaviors.
\end{itemize}

These observations raise a central question:
\begin{center}
    \textbf{ How can we suppress undesirable patterns and promote beneficial ones \\ when they cannot be reliably distinguished by terminal verifiable rewards? }
\end{center}
To address this challenge, we propose \textbf{rubrics-based GRM filtering} for collecting high-quality trajectories for Reinforced Fine-Tuning (RFT) of LLM agents.
As illustrated in Fig.~\ref{fig:workflow}, we independently sample multiple candidate continuations---either one-step actions or multistep trajectory segments---and then apply a rubrics-guided Generative Reward Model (GRM) to select the best one to proceed.
Here the rubrics are human-designed criteria that explicitly encourage or discourage specific behavioral patterns (e.g., ``create and run a reproduction script'' or ``avoid redundant tool calls'').
This introduces dense and pattern-aware feedback during trajectory rollout: whenever available among the candidates, desirable patterns are promoted, and undesired behaviors get filtered out.
Consequently, the collected dataset is enriched for high-quality task-solving actions and is diluted of inefficient or erroneous patterns, thereby providing stronger learning signals in shaping agent's behaviors than binary verifiable rewards alone.
Moreover, by explicitly encouraging generalizable behaviors (e.g. create and run reproduction scripts or unit tests), the approach supports more robust cross-domain problem-solving strategies and has potential to improve generalization performance.

We validate our method by fine-tuning LLM agents on open-source SWE benchmarks, using the seed 36B \citep{seed2025seedoss} model as the backbone and OpenHands \citep{wang2024openhands} as the agent scaffold.
Compared with naive rejection sampling that retains trajectories with maximal verifiable rewards, rubric-based GRM filtering achieves higher test accuracy and more effectively suppresses undesirable patterns while amplifying the beneficial ones, as confirmed by case analyses.
More broadly, we believe our approach can generalize and be applied to a wide range of other LLM agent applications where verifiable rewards alone are insufficient to shape high-quality intermediate behaviors.
\section{Related Works}

\textbf{LLM Coding Agents and SWE Benchmarks}~
In recent years, LLMs have been widely applied to Software Engineering \citep{liu2024large}, showing promising performance in program generation \citep{zhang2024codeagent, yeticstiren2023evaluating}, software testing \citep{wang2024software} and debugging \citep{xia2023automated}.
The emergence of LLM-based agents marks a paradigm shift from treating language models as passive sequence generators to positioning them as interactive agents that can plan, act, and learn in complex and dynamic environments.

In the domain of software engineering, \citet{jimenez2023swe} introduce SWE-Bench, a benchmark built from real GitHub issues and unit tests, and reveal that early LLMs solve only a small fraction of issue-fixing tasks.
This motivates the development of agentic systems capable of reading, editing, running, and verifying code at repository scale.
The benchmarks are enriched by a line of subsequent works \citep{pan2412training,yang2025swe,badertdinov2025swe}, while parallel efforts focus on improving agent-computer interface design and tool-use scaffolding.
SWE-agent \citep{yang2024swe} demonstrated that purpose-built interfaces (e.g. file navigation, editor actions, shell execution, and test running) can substantially improve autonomous bug-fixing and feature implementation.
In parallel, OpenHands (formerly OpenDevin) \citep{wang2024openhands} introduces an open platform where agents operate like developers (code editing, CLI use, web browsing), supporting evaluation of agents over SWE-Bench.
Complementary approaches, such as AutoCodeRover \citep{zhang2024autocoderover}, incorporate program analysis tools (AST-aware search and spectrum-based fault localization), reporting strong gains on SWE-bench-lite.

\textbf{Fine-Tuning LLMs with Verifiable Rewards}~
A growing line of research investigates fine-tuning LLMs with verifiable rewards, where supervision comes from automatically checkable signals rather than subjective human preferences.
For example, in code generation, execution-based rewards can be computed by verifying the pass of all unit tests \citep{chen2021evaluating, le2022coderl}. Similarly, mathematical reasoning tasks lend themselves to verifiable supervision, as final answers can be automatically checked against ground truth \citep{cobbe2021training}. 
Building on this idea, DeepSeek-Math \citep{shao2024deepseekmath} leverage symbolic solvers or answer checking as reward signals during fine-tuning. 
In the SWE setting, \citep{pan2412training} reports the performance improvements from fine-tuning LLM with verifiable rewards.
Despite these advances, how to move beyond sparse terminal rewards---especially in long-horizon agentic RL---remains an under-explored topic.

\textbf{Generative Reward Models and Rubrics-Guided Reward Modeling}~
Reliable evaluation is critical in LLM fine-tuning.
The concept of using LLMs for feedbacks or evaluation, also known as RLAIF \citep{bai2022constitutional} or ``LLM-as-Judge'' \citep{gu2024survey}, has emerged as a compelling approach in these years.
Unlike the traditional reward models that directly predict numerical
continuous-valued scores \citep{cobbe2021training,yu2023ovm}, the LLM evaluator, also called Generative Reward Models (GRM) or Generative Verifiers \citep{yuan2024self, mahan2024generative, chen2025rm, li2023generative, kim2024prometheus}, utilize a language model to produce natural-language judgments from which scores are derived, enabling structured reasoning and additional inference-time compute for better performance.

Human-defined rubrics or principles have been widely employed for evaluating LLM on open-ended benchmarks \citep{hashemi2024llm, arora2025healthbench, pathak2025rubric}, and for reward modeling approaches \citep{yu2025rewardanything,sun2023salmon,liu2025inferencetimescalinggeneralistreward}.
Recent work \citep{gunjal2025rubrics} further demonstrates that fine-tuning LLMs with checklist-style rubrics can yield strong performance across domains, in the absence of verifiable rewards.
In contrast, we study leveraging rubric-based reward modeling as a mechanism for trajectory filtering in SWE agents. 
Rather than replacing verifiable rewards, our goal is to complement them by shaping intermediate behaviors that terminal unit-test signals cannot reliably distinguish.
\section{Preliminary: A Rigorous Formulation of the Multi-Turn Agentic LLM}
The interaction between an LLM agent and the environment can be modeled as a finite-horizon Markov Decision Process (MDP) $M=\{\cS,\cO,\cA,T,\mP,r\}$.
Here $\cA$ is the \textit{action space}, consisting of all possible responses that invoke one of the available tools, and $\cO$ is the \textit{observation space}, containing all possible environmental feedbacks after executing an action.
We define the state as the agent's full interaction context, i.e., the prompt plus the history of tool calls and their outputs so far.
Formally, the \textit{state space} is $\cS := \cS_1\cup\cS_2...\cup\cS_T\cup\cS_{T+1}$, where $\cS_t := \cS_1\times(\cA\times\cO)^{t-1}$ represents the set of all possible context at step $t$, comprising the initial state (prompt) and the sequence of past actions and corresponding observations up to step $t$.
Besides, $T$ denotes the maximal horizon length, $\mP$ denotes the deterministic state transition function induced by context concatenation, and $r:\cS_{T+1}\rightarrow\{0,1\}$ denotes the binary terminal verifiable reward function.

The LLM agent is modeled as a policy $\pi:\cS\rightarrow \Delta(\cA)$, mapping each state to a distribution over actions.
Starting from the initial state $s_1$, at each time step (or turn) $t$, the agent samples an action $a_t \sim \pi(\cdot|s_t) \in\cA$, executes the corresponding tool call, and receives an observation feedback $o_t \in \cO$.
The next state is deterministically updated according to $s_{t+1} := s_t \circ a_t \circ o_t$, where $\circ$ denotes the token concatenation.
When the maximal horizon $T$ is reached, the agent receives a terminal score based on the final state $s_{T+1} = s_T \circ a_T \circ o_T$.
We denote the full interaction history as a \textit{trajectory} $\tau:=(s_1,a_1,o_1,a_2,o_2,...,a_T,o_T)$.
In the case that the trajectory ends early at some $t<T$ (e.g., the agent calls the ``finish'' action earlier to terminate the resolution process), we append the trajectory with null actions and observations till step $T$.
\section{Rubrics-based GRM for High-Quality RFT Data Collection}\label{sec:method}
In this section, we present the details of our method: \textbf{rubrics-based GRM filtering}.
As demonstrated in Fig.~\ref{fig:workflow}, the main idea is to incorporate rejection sampling via GRM into the trajectory generation process.
Depending on the frequency of querying GRM, we consider two variants of filtering strategy, which we refer to as Turn-Level GRM and Segment-Level GRM, respectively. 
Pseudocode for both variants is provided in Alg.~\ref{alg:turn_level_grm} and Alg.~\ref{alg:seg_level_grm}.

\textbf{GRM Filtering Strategy: Turn-Level and Segment-Level Variants}~
In the turn-level variant, the agent independently samples $N$ candidate actions at each step, where each action corresponds to a single tool call. The GRM then evaluates these candidates according to the rubrics and selects the most promising one. 
The selected action is then executed, and the remaining candidates are discarded. 
After receiving the execution results, the agent proceeds to the next step, and this process repeats until the trajectory terminates.
Finally, we add the trajectory to the RFT dataset if and only if the terminal verifiable reward is 1; otherwise, we reject it.
\begin{algorithm}
    \textbf{Input}: LLM Agent $\pi$; Candidate size $N$; Rubrics and other side-info (e.g. ground-truth git patch) $\cR$; \\
    Agent receive initial state (prompt) $s_1$.\\
    \For{$t = 1,2,...,T$}{
        $\{a_t^n\}_{n=1}^N \sim \pi(\cdot|s_t)$ \blue{\tcp*{Sample $N$ action candidates independently}} 
        $a_t \gets \text{GRM}(\{a_t^n\}_{n=1}^N, s_t, \cR)$ \blue{\tcp*{Select an action via GRM}} 
        Execute $a_t$ and get env feedback $o_{t}$.\\
        State transition: $s_{t+1} \gets s_t \circ a_t \circ o_t$.
    }
    \blue{\tcp{Accept the trajectory iff terminal reward is 1.}} 
    \lIf{$r(s_{T+1}) = 1$}{
        \textbf{Accept} $\tau := (s_1,a_1,o_1,a_2,o_2,...,a_T,o_T)$
    }
    \caption{Turn-Level Rubric-based GRM-Guided Trajectory Sampling}\label{alg:turn_level_grm}
\end{algorithm}

In the segment-level variant, we query the GRM in less frequency by evaluating and filtering multi-step trajectory segments instead---a sequence of actions along with their execution results.
Given a filtering interval $L$ (a hyperparameter), the trajectory is uniformly partitioned into $\lceil T/L \rceil$ blocks.
At the start of block $b$, from the same context state $s_{bL+1}$, we independently sample multiple continuation of length $L$. 
The GRM then evaluates these candidates, selects the best one, and discards the rest.

For the final segment $b = \lceil T/L \rceil - 1$, when the maximal horizon $T$ is reached, we can directly query the terminal verifiable reward for the entire trajectory. In practice, we simply select the trajectory with the highest reward (or a random one if all rewards are 0), instead of performing an additional GRM-based filtering step.

Roughly speaking, the turn-level version can be interpreted as a special case of the segment-level strategy with $L=1$, with one important difference: turn-level GRM selection does \emph{not} condition on execution results.
This design choice is made for practical reasons: fetching execution results for all action requires maintaining multiple concurrent environment states, which becomes computationally expensive when filtration occurs at every turn.

\begin{algorithm}
    \textbf{Input}: LLM Agent $\pi$; Candidate size $N$; Segment size $L$; Rubrics and other side-info (e.g. ground-truth git patch) $\cR$;  \\
    Agent receive initial state (prompt) $s_1$.\\
    \For{$b = 0,1,...,\lceil T/L \rceil$ - 1}{
        \For{$n=1,2,...,N$}{
            \blue{\tcp{Collect $N$ continuation starting from $s^n_{bL+1}$ for up to $L$ steps}}
            $s^n_{bL+1} \gets \text{Copy}(s_{bL+1})$ \\
            \For{$l=1,2,...,\min\{L, T - bL\}$}{
                $a^n_{bL + l} \gets \pi(\cdot|s^n_{bL+1})$ \\
                Execute $a^n_{bL + l}$ and get env feedback $o^n_{bL+l}$ \\
                $s^n_{bL+l+1}\gets s^n_{bL+l} \circ a^n_{bL + l} \circ o^n_{bL + l}$.
            }
        }
        \blue{\tcp{Select one segment via GRM}} 
        $(a_{bL+1}, a_{bL+2}, ...,a_{bL+L}) \gets \text{GRM}(\{a^n_{bL + 1} \circ o^n_{bL + 1}...,a^n_{bL + L} \circ o^n_{bL + L}\}_{n=1}^N, s_{bL+1}, \cR)$.\\
        Execute $(a_{bL+1},...,a_{bL+L})$ sequentially and get env feedbacks $(o_{bL+1},...,o_{bL+L})$.\\
        State transition: $s_{bL+L+1}\gets s_{bL+1}\circ a_{bL+1} \circ o_{bL+1},...,a_{bL+L} \circ o_{bL+L}$.
    }
    \blue{\tcp{Accept the generated trajectory iff terminal score is 1.}} 
    \lIf{$r(s_{T+1}) = 1$}{
        \textbf{Accept} $\tau := (s_1,a_1,o_1,a_2,o_2,...,a_T,o_T)$
    }
    \caption{Segment-Level Rubric-based GRM-Guided Trajectory Sampling}\label{alg:seg_level_grm}
\end{algorithm}

\textbf{GRM Evaluation Process}~
Both turn-level and segment-level filtering use the same high-level GRM prompt structure, consisting of:
\begin{enumerate}[leftmargin=*]
    \item The initial system prompt and user instruction for the agent (tool descriptions, suggested workflow, and task statement).
    \item The ground-truth git patch to fix the issue.
    \item The agent-environment interaction history (i.e., $s_t$ in Alg.~\ref{alg:turn_level_grm} or $s_{bL+1}$ in Alg.~\ref{alg:seg_level_grm}).
    \item Human designed rubrics, each with an associated weight (weights sum up to 1).
    \item The candidate actions or segments to be evaluated.
\end{enumerate}
Compared with the turn-level GRM, the segment-level prompting differs slightly to reflect longer-horizon judgements. 
We refer readers to Appx.~\ref{appx:grm_pe} for full prompt templates.

During the GRM scoring process, for turn-level setup, we instruct the GRM to score each candidate against every rubric and select the one with the highest weighted aggregate score.
For the segment-level setup, to control the context length, we adopt a pairwise comparison strategy: candidates are compared in pairs, and we select the one with the highest win rate across comparisons.

\textbf{Turn-Level v.s. Segment-Level: When to use Which?}
We now discuss the relative advantages and disadvantages of the two filtration strategies, and under which conditions each is preferable.

In general, turn-level filtration tends to yield better performance when the GRM is strong enough to accurately interpret the rubric criteria and discriminate among single-step actions given the current context.
However, if the GRM is relatively weak, especially when the action-quality differences are subtle, single-step candidates may be difficult to rank.
This is common in long-horizon agentic settings, where many local decisions have only limited downstream impact (e.g., slightly faster progress, or a small change in failure probability).

In such cases, segment-level filtration can be more effective. The multi-step trajectory segments present amplified differences in downstream utility via accumulated stochasticity across steps, producing higher-variance candidate quality and making selection easier for the GRM.
The downside is lower filtering frequency: undesirable behaviors may persist within a segment before the next selection point.
In another word, segment-level GRM inherently trades off filtering frequency for evaluation accuracy.
Overall, which strategy yields better performance therefore depends heavily on the strength and reliability of the GRM itself, and thus may vary across tasks and domains.

\section{Experiment}

In this section, we present and interpret our experiment results.
Due to the space limit, we defer additional results to Appx.~\ref{appx:experiment}.

\subsection{Setup}\label{sec:36b_setup}
\paragraph{Base Model and GRM}
All RFT experiments are conducted on \texttt{seed-OSS-36B}, which we refer to as the \emph{base model} hereafter.
We start from a checkpoint that has undergone supervised fine-tuning (SFT) on approximately 10k examples after pre-training, endowing the model with basic instruction-following and reasoning capability.
We employ \texttt{seed-1.6-thinking} as the GRM, an enhanced version of seed-1.5-thinking \citep{seed2025seed1}.
Neither the base model nor the GRM has been fine-tuned on OpenHands trajectories or SWE-specific agentic data; however, both are trained on broad coding, math, and reasoning corpora.

\paragraph{Train/Test Data}
We use open-source benchmarks for both training and testing.
The training set comprises 1.2k tasks sampled from SWE-Gym \citep{pan2412training} (200 tasks), SWE-ReBench \citep{badertdinov2025swe} (400 tasks), and SWE-Smith \citep{yang2025swe} (400 tasks).
For the test set, we adopt SWE-bench Verified, a curated subset of 500 human-validated samples from the SWE-bench test split \citep{jimenez2023swe}.

\paragraph{The OpenHands Environment}
The tools available to the agent ifall into the following categories: 
(1) basic command-line operations, such as \texttt{cd}, \texttt{ls}, \texttt{find}, \texttt{grep}, and \texttt{python}; 
(2) file-viewing actions that display the contents of a specified file within a given line range;
(3) file-editing actions that apply string replacements given a file name, an old string, and a new string;
(4) a \texttt{finish} action that terminates the trajectory.
For computational efficiency, we set the maximal horizon length t $T=20$ when collecting trajectories for training, while using a larger value $T=50$ for evaluation on the test set.
The agent may also terminate an episode early by explicitly invoking the \texttt{finish} action.

\paragraph{RFT Pipeline}
We first collect rollout trajectories from the base model on the training tasks, where we keep the RFT dataset size to 500 for consistency.
To ensure the dataset focuses on non-trivial tasks, we remove ``easy'' tasks on which our base model can achieve 100\% accuracy across five independent trials.
Subsequently, we perform RFT with the collected data, and evaluate the resulting model on the test set.

\paragraph{GRM Rubrics Design}
We define four rubrics, each assigned weight 0.25 in the overall GRM score. We list the brief descriptions here and refer the readers to Appx.~\ref{appx:grm_pe} for full rubric definitions.
\begin{itemize}[leftmargin=*]
    \item \textbf{Rubric 1: Alignment with Workflow.} We specify a step-by-step workflow highlighting key steps in the resolution process, including running tests, reproduction scripts, etc. This rubric measures how well the agent adheres to the prescribed workflow and prioritizes missing steps.
    \item \textbf{Rubric 2: Information Gain.} This rubric assesses the efficiency and informational value of the agent's actions. Higher scores are assigned to actions that advance the resolution process or may result in more useful evidence.
    \item \textbf{Rubric 3: Strategic View.} This rubric evaluates the agent's ability to synthesize prior information, reason over it, retrieve critical details to plan subsequent steps.
    \item \textbf{Rubric 4: Error Control.} This rubric penalizes unforced errors, syntax mistakes, and unproductive loops.
\end{itemize}

\subsection{Experiment Results}\label{sec:36b_result}
\textbf{Success Rate on Test Set}~
Table~\ref{tab:test_acc_weak_model} reports the test accuracy. 
Here the baseline method collects RFT data using verifiable rewards (VR) only, corresponding to the standard rejection sampling.
We compare it against four instantiations of our rubric-based GRM approaches, differing in filtering frequency (turn-level vs. segment-level) and candidate pool size.

As we can see, incorporating GRM into the trajectory collection process consistently improves test accuracy.
For the turn-level GRM (T-GRM as a shorthand), we observe a clear performance gain as the candidate size $N$ increases---an expected trend since a larger $N$ raises the likelihood of including higher-quality candidates for selection.
For a fixed candidate size $N=3$, the segment-level GRM (S-GRM as a shorthand) slightly outperforms the turn-level variant.
As discussed in Sec.~\ref{sec:method}, this can be attributed to the easier discrimination among multi-step trajectory segments, although such relative advantage may vary depending on GRM quality and the specific task.
\begin{table}[h]
    \centering
    \begin{tabular}{c|c|c|c|c|c|c}
    \hline
        & \makecell{Before\\RFT} & \makecell{Baseline \\ (VR only)} & \makecell{T-GRM\\$N=3$} & \makecell{T-GRM\\$N=5$} & \makecell{S-GRM\\$L=7, N=3$} & \makecell{S-GRM\\$L=5,N=3$} \\ \hline
      Test Accuracy  &  28.0\% & 36.6\% & 39.3\% & 42.6\% & 40.2\% & 40.5\%\\ \hline
    \end{tabular}
    \caption{\textbf{Test Accuracy of Models after RFT.} Collecting RFT dataset with our rubric-based GRM approach leads to higher test accuracy.}\label{tab:test_acc_weak_model}
\end{table}

\textbf{Pattern Analysis after RFT on Test Set}~
Beyond test accuracy, Table~\ref{tab:case_analysese_test_set} presents the pattern analysis to verify whether our rubrics-based GRM filtering effectively promotes desired patterns while suppressing undesired ones. 
We examine three aspects: (i) creation/execution of tests or reproduction scripts, (ii) error control, and (iii) resolution efficiency.
For each test instance, we sample one trajectory per test task, and compute the ratio of exhibiting the corresponding behaviors.
In Appx.~\ref{appx:experiment}, we also provide the pattern analysis on the collected training data.

Table~\ref{tab:case_analysese_test_set}-(a) reports the proportion of trajectories containing test or reproduction related behaviors, identified via keyword-based regex matching.
Our approach achieves a noticeably higher ratio of such desired patterns, indicating its effectiveness compared with filtering based solely on verifiable rewards.

Next, we analyze execution errors, focusing on ``self-correctable'' and avoidable cases---those most relevant to our motivation of this paper---including: (a) accessing non-existent paths; (b) invalid view ranges when inspecting files (e.g., start line exceeds end line); (c) string replacement failures in file edition (e.g., strings to replace are invalid or unfounded).
Our method yields a clear reduction in such errors compared with the baseline, demonstrating the role of GRM filtering in improving reliability and robustness.

Lastly, we report the average number of turns per episode as a proxy for the efficiency of the resolution process.
The intervention of rubric-based GRM notably improves efficiency on the test set.
Besides the reduction of error rate discussed before, we attribute this improvement primarily to Rubric 2 and Rubric 3.
Rubric 2 encourages actions with higher information gain during RFT data collection---for example, we observe GRM prefers ``\texttt{viewing the first 400 lines}'' to ``\texttt{viewing the first 200 lines}'' of the same files; or prefers a combined consecutive command lines ``\texttt{cd ... \&\& grep/find ...}'' over simply ``\texttt{cd ...}''.
Meanwhile, given the access to the ground-truth git patch, which indicates the ground-truth files or lines of codes to edit, Rubric 3 encourages GRM to identify and prioritize actions that meaningfully advance the resolution process over peripheral or cosmetic ones.
Consequently, the filtered dataset contains a higher concentration of efficient and purposeful actions, leading the fine-tuned model to exhibit more effective and goal-directed trajectories.

\begin{table}[h]
    \centering
    \begin{subtable}{1.0\linewidth}
    \begin{tabular}{c|c|c|c|c|c|c}
    \hline
        & \makecell{Before\\RFT} & \makecell{Baseline \\ (VR only)} & \makecell{T-GRM\\$N=3$} & \makecell{T-GRM\\$N=5$} & \makecell{S-GRM\\$L=7, N=3$} & \makecell{S-GRM\\$L=5, N=3$} \\ \hline
        Create Test & 74.6\% & 80.5\% & 82.1\% & 79.6\% & 78.6\% & \textbf{84.7\%} \\
        Create Repro. Script  &  3\% & 6.6\% & 8.4\% & \textbf{9.8\%} & 6.6\% & 8.4\% \\
        Run Test & 76.6\% & 81.7\% & 86.9\% & 87.8\% & \textbf{89.6\%} & 87.1\% \\
        Run Repro. Script & 3\% & 6.6\% & 8.4\% & \textbf{9.8\%} & 6.6\% & 8.4\% \\\hline
    \end{tabular}
    \caption{Create and Run Test/Reproduction Behaviors}
    \label{tab:run_test_repro_36b}
    \end{subtable}
    \begin{subtable}{1.0\linewidth}
    \centering
        \begin{tabular}{c|c|c|c|c|c}
        \hline
            & \makecell{Baseline \\ (VR only)} & \makecell{T-GRM\\$N=3$} & \makecell{T-GRM\\$N=5$} & \makecell{S-GRM\\$L=7, N=3$} & \makecell{S-GRM\\$L=5, N=3$} \\ \hline
        Error Rate (Task-Level) & 32.7\% & 31.3\% & 27.8\% & 28.2\% & \textbf{27.4\%}\\
        Error Rate (Turn-Level) & 3.4\% & 3.1\% & 2.9\% & 3.3\% & \textbf{2.4\%} \\\hline
        \# Avg. Turn   &  35.2 & 34.3 & \textbf{32.0} & 33.8 & 33.8\\
        \hline
        \end{tabular}
        \caption{Resolution Efficiency and Error Control}
        \label{tab:efficiency_and_error}
    \end{subtable}
        \caption{\textbf{Pattern Analysis of 36B on Test Set after RFT.} \textbf{Up}: the proportion of resolution trajectories create or run test/reproduction scripts. \textbf{Down}: the averaged consumed turns to resolve the issue and the error rate. Here ``Task-Level'' measures the proportion of tasks (or trajectories) containing error, while ``Turn-Level'' computes the proportion of overall turns with error.
        Overall, incorporating GRM into the RFT data collection stage helps to promote desired patterns while suppressing undesired ones in LLM agents' behaviors.}
    \label{tab:case_analysese_test_set}
\end{table}

\subsection{Verification Ability is Sufficient for GRM}

A natural question is whether the GRM must be stronger than the base model in order to provide useful guidance during trajectory filtering.
To investigate this, we first note that our GRM achieves 43\% test accuracy on SWE-bench Verified, which is higher than the 28\% by the base model reported in Sec.~\ref{sec:36b_result}.
Therefore, in this section, we examine a stricter regime: can GRM filtering still help when the base model is stronger than the GRM itself?

To obtain this stronger base model, we start from the same 36B pre-training checkpoint and perform the same SFT procedure, but augment the SFT dataset with 500 additional resolution trajectories generated by Claude Opus 4 as the LLM agent backbone.
This increases the base model's initial test accuracy to 48\%, surpassing that of the GRM.

We keep the training and test datasets, as well as the pipeline, identical to Sec.~\ref{sec:36b_result}.
The only change is to increase maximum horizon length to $T=50$ during the data collection, because the new base model typically requires more than 20 steps to complete the resolution process in most instances.

\begin{table}[h]
    \centering
    \begin{tabular}{c|c|c|c}
    \hline
        & \makecell{Before\\RFT} & \makecell{Baseline \\ (VR only)} & \makecell{T-GRM\\$N=3$}  \\ \hline
      Test Accuracy  &  48.0\% & 46.8\% &  \textbf{50.6\%} \\ \hline
    \end{tabular}
    \caption{\textbf{Test Accuracy of the Stronger 36B after RFT.} Our approach still yields a better performance. The baseline method results in a performance drop, which we interpret as a normal fluctuation when the performance is close to its upper limit.}
    \label{tab:test_score_strong_36b}
\end{table}

\begin{table}[h]
    \begin{subtable}{1.\linewidth}
    \centering
    \begin{tabular}{c|c|c|c}
    \hline
        & \makecell{Before\\RFT} & \makecell{Baseline \\ (VR only)} & \makecell{T-GRM\\$N=3$} \\ \hline
        Create Test &  94.4\% & \textbf{95.4\%} & 94.2\% \\
        Create Repro. Script &  19\% & 23.4\% & \textbf{25.6\%} \\
        Run Test & 96\% & \textbf{97\%} & 95.8\%  \\
        Run Repro. Script & 18.6\% & 23.4\% & \textbf{25.6\%}  \\\hline
    \end{tabular}
    \caption{Create and Run Test/Reproduction Behaviors}
    \label{tab:run_test_repro_strong_36b}
    \end{subtable}
    \begin{subtable}{1.0\linewidth}
    \centering
    \begin{tabular}{c|c|c}
    \hline
       & \makecell{Baseline \\ (VR only)} & \makecell{T-GRM\\$N=3$} \\ \hline
      Error Rate (Task-Level) & {15.4\%} & 15.8\%  \\
      Error Rate (Turn-Level) & {1.0\%} & 1.1\% \\
      \# Avg. Turn   &  44.8 & {41.8} \\
    \hline
    \end{tabular}
    \caption{Resolution Efficiency and Error Control}
    \label{tab:efficiency_and_error_strong_36b}
    \end{subtable}
    \caption{\textbf{Pattern Analysis of the ``stronger'' 36B on Test Set after RFT.} We observe clear improvements in reproduction-related behaviors and the averaged number of consumed turns. Although there is no significant change in the test-related behaviors or error rates, this is expected---the base model already exhibits near-optimal performance for those patterns.}\label{tab:case_analysis_strong_36b}
\end{table}

Table~\ref{tab:test_score_strong_36b} and Table~\ref{tab:case_analysis_strong_36b} summarize the results.
Our approach continues to demonstrate clear advantages in test accuracy, reproduction-related behaviors, and the average number of turns consumed, despite the GRM being weaker than the base model.
Overall, these findings suggest that the GRM need not be an expert generator, and possessing a basic verification capability is sufficient to provide valuable learning signals.
We interpret this through the lens of the generation-verification gap \citep{song2024mind,chen2025sets}: a weaker model that can reliably evaluate rubric-aligned behavior (especially with access to ground-truth side information) can still guide the learning of a stronger generative model.

\section{Conclusion}
Motivated by the limitation of verifiable rewards, we propose leveraging rubric-based generative reward models (GRMs) as an additional source of learning signals for fine-tuning LLM agents.
Our RFT experiments demonstrate that this GRM-guided data collection promotes desirable behavioral patterns while suppressing undesirable ones, leading to improved generalization on the test set.
Notably, these gains do not require an expert-level GRM---strong verification ability alone is sufficient to provide effective guidance.

Several directions remain open for future research.
Firstly, it would be valuable to extend our approach to the reinforcement learning training stage, using rubric-based GRM signals to densify sparse rewards and accelerate policy improvement.
Secondly, the design of effective rubrics remains a challenging and largely manual process; developing systematic methods for rubric construction and for identifying rubric criteria that improve generalization remains an open challenge.
Finally, we believe our approach holds promise beyond software-engineering tasks---for example, in search agents, deep research assistants, and other LLM-agent domains---and we leave such explorations to future work.

\clearpage

\bibliographystyle{plainnat}
\bibliography{references}

\clearpage

\newpage
\beginappendix

\section{Details in GRM Prompt Engineering}\label{appx:grm_pe}

\subsection{GRM Prompt Structures}
For turn-level GRM, we compare all the actions together.

\begin{tcolorbox}[promptbox, title=GRM Prompt Structure for Turn-Level Filtering]\small
    You are a professional evaluator for a GitHub issue resolution task. In this scenario, a large language model (LLM) assistant is given a GitHub issue and instructed to resolve it utilizing a set of available tools (function calls). At each step of the process, the LLM assistant generates an action (function call), the user executes it, and returns the execution results. The LLM leverages these results in the subsequent steps in the resolution process.
    \\
    
    \textbf{Your role as an evaluation expert} \\
    You will be provided with:
    \begin{itemize}
        \item "len(responses)" candidate actions generated by the LLM at a specific stage of the resolution process;
        \item The full conversation history between the user and the LLM agent (to provide context and track completed steps);
        \item Supplementary references, including the ground-truth git patch (the validated solution to the issue).
    \end{itemize}
    
    Based on these information, your task is to evaluate all {len(responses)} candidate actions and select the single best one to execute. Only your selected action will be executed; all the others will be discarded.
    \\
    
    \textbf{Step-by-Step Evaluation Instructions}
    \\    
    \underline{Step 1: Deeply Understand the Context (Preparation)}

    Before scoring, you must first become an expert on the problem.
    \begin{enumerate}
        \item Analyze the User Instruction: Read the GitHub issue carefully. Identify the reported bug or requested feature and the exact solution requirements.
        \item Study the Ground-Truth Patch: Examine the validated solution. Note which files changed and the nature of the change (for example, logic correction, wrong variable, missing condition). Use this to judge whether a segment is moving toward the real solution.
        \item Review the Conversation History: Read through the sequence of actions taken by the LLM and the results it received. Get a sense of the model's approach so far.
    \end{enumerate}

    \underline{Step 2: Score the Candidate Actions by the Rubrics}
    
    Carefully review the user's question and the conversation context to understand both the task and the current resolution stage. Then, evaluate each candidate action based on the following rubrics (0-4 points for each item):\\

    \begin{center}
        \texttt{PlaceHolder for Rubrics}
    \end{center}
    \vspace{1em}
    
    \underline{Step 3: Evaluate One by One and Output Conclusion}
    \begin{itemize}
        \item 
    Score each candidate action according to the above rubrics and briefly explain the reasons. 
        \item 
    Report the weighted total score = sum(score * corresponding weight percentage) for each candidate action. Compare the weighted total scores of all actions to determine the best-performing one.
        \item 
    Write the conclusion strictly in the following format:
    \begin{center}
        ACTION i WINS
    \end{center}
    where i is a number between 1 and {len(responses)}, corresponding to the chosen optimal action.
    \end{itemize}
\end{tcolorbox}

\newpage

For segment-level GRM, due to the limit of context length, we instead enumerate all the possible pairs of candidates and conduct pairwise comparison.
Slightly different from turn-level version, here we ask GRM analyize and output the better one without scoring each candidate.

\begin{tcolorbox}[promptbox, title=GRM Prompt Structure for Segment-Level Filtering]\small
    You are a professional evaluator for a GitHub issue resolution task. In this scenario, a large language model (LLM) assistant is given a GitHub issue and instructed to resolve it utilizing a set of available tools (function calls). At each step of the process, the LLM assistant generates an action (function call), the user executes it, and returns the execution results. The LLM leverages these results in the subsequent steps in the resolution process.
    \\
    
    \textbf{Your role as an evaluation expert} \\
    You will be provided with:
    \begin{itemize}
        \item Two trajectories to evaluate ("Trajectory" is a sequence of actions and their execution results generated by the LLM at an intermediate stage of the resolution process);
        \item The full conversation history between the user and the LLM agent (to provide context and track completed steps);
        \item Supplementary references, including the ground-truth git patch (the validated solution to the issue).
    \end{itemize}
    
    Based on these information, your task is to evaluate all {len(responses)} candidate actions and select the single best one to execute. Only your selected action will be executed; all the others will be discarded.
    \\
    
    \textbf{Step-by-Step Evaluation Instructions}
    \\    
    \underline{Step 1: Deeply Understand the Context (Preparation)}

    Before scoring, you must first become an expert on the problem.
    \begin{enumerate}
        \item Analyze the User Instruction: Read the GitHub issue carefully. Identify the reported bug or requested feature and the exact solution requirements.
        \item Study the Ground-Truth Patch: Examine the validated solution. Note which files changed and the nature of the change (for example, logic correction, wrong variable, missing condition). Use this to judge whether a segment is moving toward the real solution.
        \item Review the Trajectories: Read through the sequence of actions taken by the two trajectories and the results they received. Get a sense of the model's approach so far.
    \end{enumerate}

    \underline{Step 2: Score the Candidate Actions by the Rubrics}
    
    Carefully review the user's question and the conversation context to understand both the task and the current resolution stage. Then, evaluate each candidate trajectory segments based on the following rubrics (0-4 points for each item):\\

    \begin{center}
        \texttt{PlaceHolder for Rubrics}
    \end{center}
    \vspace{1em}
    
    \underline{Step 3: Output Your Evaluation Result}
    \begin{itemize}
        \item Based on your evaluation result, select the better one between the two provided trajectories.
        \item Output "YES" if you believe the first trajectory is better, and "NO" if you believe the second trajectory is better.
    \end{itemize}
\end{tcolorbox}

\newpage
\subsection{Details of Rubrics Structures}

\begin{tcolorbox}[rubricsbox, title=Rubrics for Turn-Level Filtering]\small
\textbf{Rubric 1: Alignment with the Ideal Workflow} (Weight: 25\%)

The ideal workflow of the resolution process should, in sequence, incorporate the following key steps:
\begin{enumerate}
    \item Running existing tests in the repository.
    \item Identify and inspect the files relevant to the problem and its solution.
    \item Create and execute a reproduction script (e.g., `reproduce\_error.py`) to recreate the error or problematic state, if feasible.
    \item Edit relevant files to fix the bug or implement the required change.
    \item Re-run the repository's existing test cases and the reproduction script to confirm the issue is solved.
    \item Develop and execute a more comprehensive test script (e.g., `comprehensive\_tests.py`) to check for edge cases. 
    \item Conclude with a summary of the changes and finalize the resolution.
\end{enumerate}

How well does the action align with the ideal workflow above, addresses uncompleted critical steps, and focuses on the core of the user's question and the task objectives?
Higher scores for actions that advance new and essential steps in the workflow that have not yet been completed, and for focusing on elements central to the user’s question and the validated solution.
Especially, strongly prioritize \underline{running tests} and \underline{creating/running a reproduction script} if not yet done.\\

\textbf{Rubric 2. Information Gain and Progress} (Weight: 25\%)

What is the expected value or utility of the information produced by this action, given the information collected so far in the conversation context? 
Higher scores for actions likely to help identify root causes or correctly implement a solution.\\

\textbf{Rubric 3. Strategic Value} (Weight: 25\%)

How well does the action advance the resolution process through strategic decision-making? 
This includes abilities such as integrating prior information, retrieving critical details, and driving meaningful progress toward solving the issue (e.g., narrowing down possibilities, eliminating dead ends).\\

\textbf{Rubric 4. Error/Noise Control} (Weight: 25\%)

To what extent does the action avoid execution error, noise or redundancy? Lower scores for:
\begin{itemize}
    \item Actions contain incorrect syntax or are likely to result in execution error;
    \item Repeating the same failed actions without learning, (e.g., applying the same failed `str\_replace` multiple times)
    \item Irrelevant, misleading, or low-value actions.
\end{itemize}
\end{tcolorbox}

\newpage
\begin{tcolorbox}[rubricsbox, title=Rubrics for Segment-Level Filtering]\small
\textbf{Rubric 1: Alignment with the Ideal Workflow} (Weight: 25\%)

The ideal workflow of the resolution process should, in sequence, incorporate the following key steps:
\begin{enumerate}
    \item Running existing tests in the repository.
    \item Identify and inspect the files relevant to the problem and its solution.
    \item Create and execute a reproduction script (e.g., `reproduce\_error.py`) to recreate the error or problematic state, if feasible.
    \item Edit relevant files to fix the bug or implement the required change.
    \item Re-run the repository's existing test cases and the reproduction script to confirm the issue is solved.
    \item Develop and execute a more comprehensive test script (e.g., `comprehensive\_tests.py`) to check for edge cases. 
    \item Conclude with a summary of the changes and finalize the resolution.
\end{enumerate}

How well does the trajectory align with the ideal workflow above, addresses uncompleted critical steps, and focuses on the core of the user's question and the task objectives?
\begin{itemize}
    \item A good trajectory should demonstrate a masterful adherence to the workflow. It prioritizes running tests to establish a baseline, intelligently inspects files directly related to the ground-truth patch, and makes a clear, logical progression toward verification.
    \item A bad trajectory may show little regard for a structured workflow. It may perform random file inspections, ignore failing test results, or take actions that are out of sequence and unproductive.
\end{itemize}
\vspace{1em}

\textbf{Rubric 2. Information Gain and Progress} (Weight: 25\%)

This rubric assesses the efficiency and value of the LLM's actions. Is each step purposeful and does it move the process closer to a solution?

\begin{itemize}
    \item A good trajectory should include a majority of actions that are productive and contribute to forward progress. There might be one or two minor detours or less impactful actions, but the overall trajectory is one of clear advancement.
    \item A bad trajectory may generate no useful information or actively move the process backward. The trajectory is entirely composed of irrelevant, misguided, or counter-productive steps.
\end{itemize}
\vspace{1em}

\textbf{Rubric 3. Strategic Value} (Weight: 25\%)

This rubric evaluates the LLM's higher-level reasoning, including abilities such as integrating prior information, retrieving critical details, and driving meaningful progress toward solving the issue (e.g., narrowing down possibilities, eliminating dead ends). Does it connect information across turns, form logical hypotheses, and adapt its strategy based on new information?

\begin{itemize}
    \item A good trajectory should show good strategic decision-making. The LLM effectively uses prior context and results to guide its actions. Its reasoning is sound, logical, and advances the solution.
    \item A bad trajectory usually shows no evidence of a strategy. Actions appear random and disconnected from the problem context and the information gathered so far.
\end{itemize}
\vspace{1em}

\textbf{Rubric 4. Error/Noise Control} (Weight: 25\%)

This rubric assesses to what extent the trajectory avoid unforced errors, syntax mistakes, and unproductive loops.

\begin{itemize}
    \item A good trajectory contains a minor, isolated error, such as a single typo in a command or one redundant action that doesn't significantly derail the process. The model may even self-correct in the next step.
    \item A bad trajectory is plagued by errors or inefficiencies. It contains multiple syntax errors, repeats the same failed actions without learning, or diverges into a long, unproductive sequence of actions that must be abandoned.
\end{itemize}
\end{tcolorbox}

\newpage

\section{Additional Experiment Results}\label{appx:experiment}

\paragraph{Pattern Analysis of 36B on Training Dataset}
Table~\ref{tab:behavioral_analysis_36b_train_set} presents the case analyses of 36B base model (Sec.~\ref{sec:36b_result}) on the training dataset.
In general, our GRM-based approach provides higher-quality datasets, as reflected by improvement on creation/execution of test and reproduction scripts, a lower error ratio and a reduced average of consumed turns.
These observations align with and help explain the model's improved behavior on the test set after the RFT stage described in Sec.~\ref{sec:36b_result}.

\begin{table}[h]
    \begin{subtable}{1.0\linewidth}
    \centering
    \begin{tabular}{c|c|c|c|c|c}
    \hline
        & \makecell{Baseline \\ (VR only)} & \makecell{T-GRM\\$N=3$} & \makecell{T-GRM\\$N=5$} & \makecell{S-GRM\\$L=7, N=3$} & \makecell{S-GRM\\$L=5, N=3$}\\ \hline
      Error Ratio (Task-Level) & 51.6\% & 48.3\% & 48.1\% & 43.6\% & \textbf{39.8\%}\\
      Error Ratio (Turn-Level) & 5.7\% & 5.7\% & 4.3\% & 4.0\% & \textbf{3.5\%}\\\hline
      Avg Turn   &  17.3 & 16.1 & \textbf{15.1} & 16.8 & 16.6\\
      \hline
    \end{tabular}
    \caption{Resolution Efficiency and Error Control}
    \label{tab:eff_err_36B_train_set}
    \end{subtable}
    \begin{subtable}{1.0\linewidth}
    \centering
    \begin{tabular}{c|c|c|c|c|c}
    \hline
        & \makecell{Baseline \\ (VR only)} & \makecell{T-GRM\\$N=3$} & \makecell{T-GRM\\$N=5$} & \makecell{S-GRM\\$L=7, N=3$} & \makecell{S-GRM\\$L=5, N=3$}\\ \hline
     Create Test   & 52.5\% & 59.7\% & 60.0\% & 59.7\% & \textbf{60.9\%}  \\
     Create Repro. Script   & 20.3\% & 21.5\% &  \textbf{21.7\%} & 19.3\% & 20.3\%  \\
     Run Test   & 64.2\% &  \textbf{74.1\%} &  73.8\% & 70.8\% &  71.4\% \\
     Run Repro. Script  & 19.9\% &  21.2\% &  \textbf{21.6\%} & 18.9\% & 20.1\% \\ \hline
    \end{tabular}
    \caption{Create and Run Test/Reproduction Behaviors}
    \label{tab:test_repro_36B_train_set}
    \end{subtable}
    \caption{\textbf{Pattern Analysis of 36B Model on Train Set}}\label{tab:behavioral_analysis_36b_train_set}
\end{table}

\end{document}